\documentclass{iopjournal}

\usepackage{amsmath,amssymb,amsfonts}
\usepackage{array}
\usepackage{float}
\usepackage{url}

\begin{document}

\articletype{Paper}

\title{An Electric Potential-Augmented Benchmark Dataset for Physics-Guided Image Reconstruction of Electrical Capacitance Tomography}

\author{Xinqi Zhang$^1$, Qiming Ma$^1$ and Lihui Peng$^{1,*}$} 

\affil{$^1$Department of Automation, Tsinghua University, Beijing 100084, China}

\affil{$^*$Author to whom any correspondence should be addressed.}

\email{lihuipeng@mail.tsinghua.edu.cn}


\begin{abstract}
While deep learning has significantly advanced image reconstruction of Electrical Capacitance Tomography (ECT), most data-driven methods map directly between capacitance and permittivity distribution, treating the sensor as a black box. This overlooks the electric potential field---the fundamental physical link governing the nonlinear and ill-posed ``soft-field'' effect. To address this, we propose an electric potential-augmented ECT benchmark dataset designed to explicitly integrate latent physics behind ECT into the learning process. Generated via a COMSOL-MATLAB pipeline for an eight-electrode sensor as an example, the dataset comprises 20,000 randomized samples across four typical flow patterns. Crucially, alongside the conventional capacitance vectors and permittivity distributions depicted as images, each sample preserves eight excitation-wise full-field potential maps. Beyond data release, we provide illustrative evaluation protocols for both forward and inverse problems of ECT. Through comprehensive testing on both in-distribution (IID) and out-of-distribution (OOD) scenarios, we systematically demonstrate how the inclusion of electric potential maps enhances modeling accuracy and robustness. Fundamentally, the explicit inclusion of latent field information significantly lowers the barrier to integrating physical laws into ECT modeling, thereby establishing a standardized foundation for future physics-guided machine learning of ECT image reconstruction.
\end{abstract}

\section{Introduction}

Electrical capacitance tomography (ECT) is a well-established soft-field imaging modality for non-intrusive monitoring and visualization of multiphase flows and related industrial processes \cite{huang1989,xie1992,reinecke1996,marashdeh2008}. Since its early development, substantial effort has also been devoted to ECT sensor design for different applications because the electrode arrangement, shielding structure, and sensing-domain geometry directly affect the measured capacitances and the achievable reconstruction quality \cite{yang2010sensor}. Unlike hard-field modalities such as X-ray CT, however, the sensitivity distribution in ECT is not fixed but depends on the unknown permittivity distribution itself. This soft-field effect makes image reconstruction strongly nonlinear and ill-posed, and it has long made reconstruction algorithm design a central issue in ECT \cite{isaksen1996,yang2003}.

Accordingly, a large body of work has focused on ECT image reconstruction methods. Classical approaches include linear iterative algorithms such as Landweber-type schemes \cite{yang1999landweber}, regularization-based and nonlinear regularized inversion \cite{peng2000regularization,fang2004,soleimani2005}, nonlinear Landweber iteration \cite{li2008nonlinearLandweber}, statistical and Bayesian formulations \cite{watzenig2009}, and sparse-reconstruction approaches in electrical tomography and ECT \cite{zhao2014sparse,ye2015sparse}. These methods remain important because they explicitly address the ill-posed nature of the ECT inverse problem and provide efficient baselines for practical imaging. Nevertheless, their reconstruction accuracy is often limited by linearization error, the limited number of independent electrode-pair capacitance measurements, measurement noise, and the difficulty of representing complicated permittivity distributions using a fixed inverse model.

Data-driven ECT image reconstruction is also not entirely new. Neural-network-based reconstruction had already been explored in the 1990s \cite{nooralahiyan1997}, and deep learning has further accelerated this direction in recent years. Representative studies include autoencoder-based reconstruction \cite{zheng2018ae}, CNN-based reconstruction \cite{zheng2019cnn}, deep-learning-compensated back-projection \cite{zheng2020dlbp}, U-Net-based reconstruction \cite{yang2019unet}, visual-representation-based deep neural networks \cite{zhu2020visual}, adversarial reconstruction and enhancement methods \cite{deabes2022adversarial}, and recently proposed machine-learned inverse operators \cite{wanta2025mlpseudoinverse}. The recent review by Peng et al. summarizes the rapid progress of deep learning-based image reconstruction for ECT and identifies data, neural-network design, and physical modeling as closely related factors in this research direction \cite{peng2025review}. This observation is particularly important for ECT because a dataset is not merely a collection of training samples. It implicitly defines the sensor geometry, excitation protocol, discretization, permittivity distribution, measurement format, output representation, and train--test setting under which a machine learning-based reconstruction method is developed and evaluated.

From this perspective, benchmark datasets play a more fundamental role in ECT than in many standard image-to-image learning tasks. Since the measured capacitances are integral quantities induced by excitation-dependent electric-field responses in a soft-field sensing process, the data modality determines which part of the electrostatic process is accessible to the learning algorithm. If a dataset only preserves capacitance vectors and permittivity maps, machine learning-based methods are naturally encouraged to approximate a direct black-box mapping from measured capacitances to permittivity distributions. Such a setting is useful and has enabled substantial progress, but it omits the internal electric potential field that physically connects the material distribution with the electrode-pair capacitance measurements. Therefore, improving the dataset formulation itself is a necessary step toward physics-guided machine learning for ECT image reconstruction.

The benchmark dataset introduced by Zheng et al. was an important milestone because it provided unified capacitance-permittivity-distribution pairs for training, testing, and comparing machine learning-based ECT image reconstruction methods under common criteria \cite{zheng2018dataset}. More recently, Shi et al. introduced ECT-Bench, a large-scale experimentally acquired benchmark for planar ECT, which provides annotated mutual-capacitance measurements collected with a custom $3\times3$ electrode array under contact, non-contact, and dynamic acquisition settings \cite{shi2026dataset}. These benchmark efforts highlight the increasing importance of open and standardized data resources for reproducible ECT reconstruction and sensing research. At the algorithmic level, deep-learning-based ECT reconstruction is still commonly formulated as paired supervised learning, where capacitance measurements are used as inputs and permittivity distributions are used as reconstruction targets \cite{peng2025review}. Meanwhile, several recent works have attempted to incorporate measurement physics more explicitly into the learning process. Jin et al. proposed a physics-constrained deep learning method for ECT image reconstruction, highlighting the value of embedding physical constraints into data-driven inversion \cite{jin2024physics}. Lei and Liu further proposed a data- and measurement-mechanism-integrated imaging method that combines data-driven modeling with the ECT measurement mechanism \cite{lei2024mechanism}. These studies show that ECT reconstruction can benefit from physical constraints and model-based learning. However, existing datasets and methods generally do not provide a reusable data modality in which excitation-wise intermediate field quantities are explicitly preserved for broader forward and inverse problems of ECT.

From an electromagnetic viewpoint, permittivity and capacitance are coupled through the internal electric potential field rather than through a direct algebraic mapping. Under the quasi-static approximation, the electric potential field $\phi$ satisfies 
\begin{equation} 
\nabla \cdot (\epsilon \nabla \phi)=0, 
\label{eq:potential_governing}
\end{equation} 
and capacitance measurements are derived from the associated electrostatic field response \cite{xie1992,reinecke1996,yang2003}. This indicates that electric-potential maps contain physically meaningful information about how the permittivity distribution distorts the electric field inside the sensing domain. 

Field information itself is not new in ECT. Loser et al. investigated image reconstruction along electric field lines by computing the internal electrostatic field using FEM, visualizing the distortion of electric field lines caused by high-permittivity regions, and incorporating field-line-based weighting matrices into ART/LBP reconstruction \cite{loser2001}. More recently, Wang et al. proposed a digital-twin-assisted three-dimensional ECT framework that integrates a 3-D fluid--electrostatic field coupling model to generate virtual multiphase-flow imaging data \cite{wang2024digitaltwin}. Banasiak et al. further investigated direct estimation of electric-field distributions in circular ECT sensors using graph convolutional networks, where FEM-simulated field data were used as supervised targets for learning excitation-dependent field responses \cite{banasiak2025gcnfield}. These studies indicate that field-level quantities, including electric potential and electric-field information, are gaining increasing attention as physically meaningful learning targets in ECT modeling. Nevertheless, their main focus is either digital-twin-assisted 3D imaging or fast surrogate modeling of the forward field solution, rather than the construction of an electric potential-augmented benchmark dataset that jointly preserves capacitance measurements, permittivity distributions, and excitation-wise full-field potential maps for systematic evaluation.

To address this gap, this paper presents an electric potential-augmented ECT benchmark dataset constructed using a reproducible COMSOL-MATLAB pipeline. The dataset contains 20,000 randomized samples across four typical flow patterns. Following the sensor configuration of Zheng et al. \cite{zheng2018dataset}, we adopt the same eight-electrode setting and further augment each sample with eight excitation-wise full-field potential maps. Thus, each sample includes the permittivity distribution, a 28-dimensional capacitance vector, and eight excitation-wise full-field potential maps. Using this dataset, we further study the role of electric potential maps in ECT modeling and image reconstruction, with particular attention to both in-distribution and out-of-distribution evaluations.

The main contributions of this paper are summarized as follows:

\begin{itemize}
    \item \textbf{Electric potential-augmented dataset:} We construct a 20,000-sample ECT benchmark dataset that extends conventional capacitance-permittivity-distribution pairs with eight excitation-wise full-field electric potential maps.

    \item \textbf{Experimental validation of electric potential maps:} We conduct systematic forward and inverse ECT experiments under both in-distribution and out-of-distribution settings, demonstrating the effectiveness of incorporating electric potential maps in ECT modeling and image reconstruction.

    \item \textbf{Support for physics-guided machine learning:} By retaining intermediate field information that is typically absent from public ECT datasets, the proposed dataset provides a practical testbed for future physics-guided machine learning methods for ECT image reconstruction.
\end{itemize}

\section{Dataset Construction and Characterization}

\subsection{Sensor Modeling and Physical Configuration}
\begin{figure}[H]
    \centering
    \includegraphics[width=0.7\textwidth]{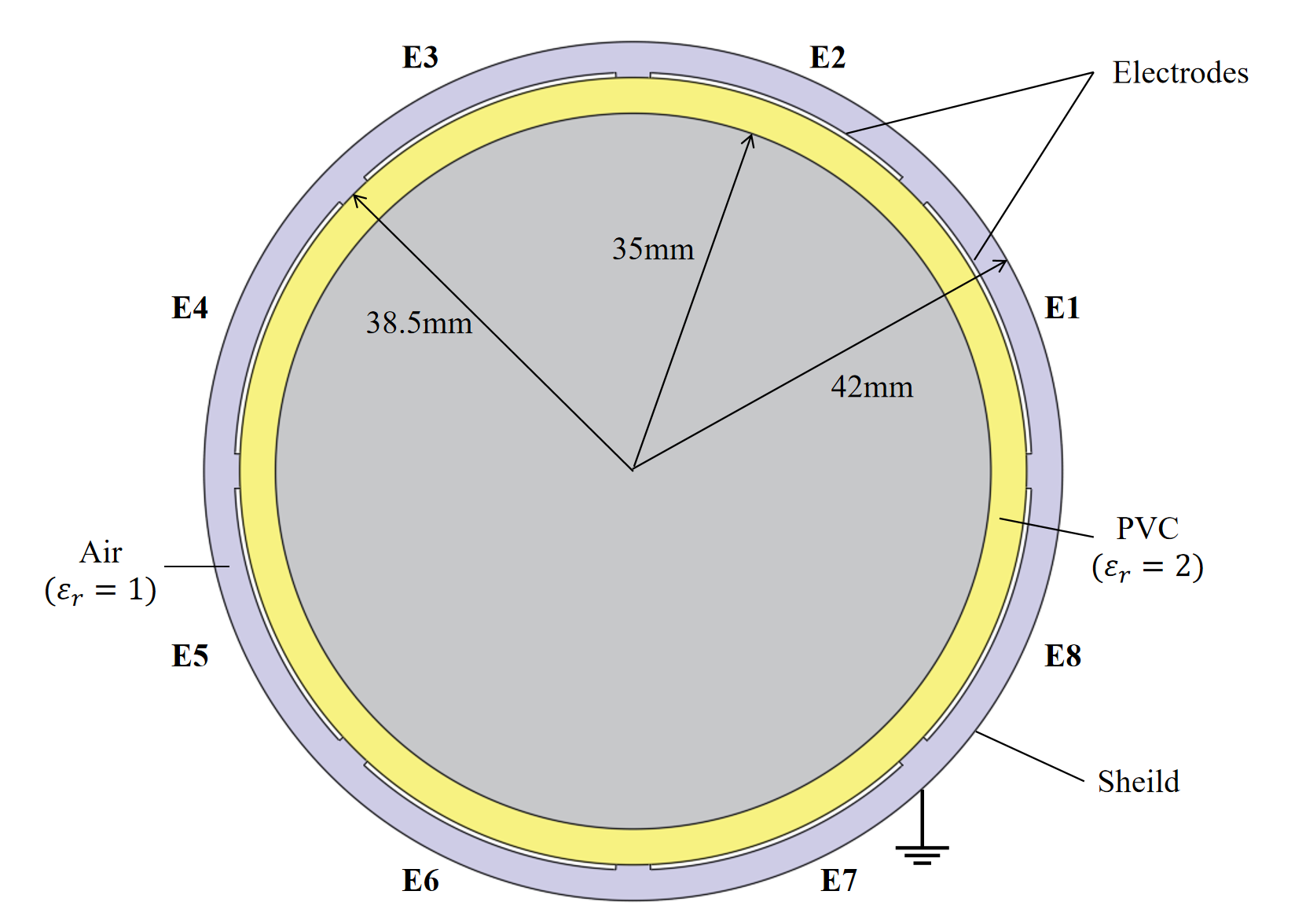}
    \caption{Geometric configuration of the eight-electrode ECT sensor.}
    \label{fig:sensor_structure}
\end{figure}

The proposed dataset is generated using a two-dimensional cross-sectional ECT sensor model with eight external measurement electrodes, as illustrated in figure~\ref{fig:sensor_structure}. The sensor adopts a conventional circular pipe configuration for multi-electrode capacitance sensing. The inner and outer radii of the insulating pipe are 35~mm and 38.5~mm, respectively, and the radius of the outer shield is 42~mm. The eight electrodes are uniformly arranged around the pipe circumference, with each electrode subtending an angular span of 40$^\circ$ and leaving an inter-electrode gap of 5$^\circ$.

Based on this fixed sensor geometry, the numerical model assigns material properties to both the imaging domain and the surrounding regions. The relative permittivities of the low- and high-permittivity phases are set to $\epsilon_{\mathrm{low}}=1.0$ and $\epsilon_{\mathrm{high}}=2.7$, respectively, while the PVC pipe wall and the exterior air region are assigned relative permittivities of 2.0 and 1.0. These settings provide a representative dielectric contrast for two-phase ECT imaging. Under this configuration, changes in the cross-sectional permittivity distribution give rise to corresponding variations in the boundary capacitance measurements. Consistent with the soft-field effect of ECT, the sensing field is globally affected by the permittivity distribution within the imaging domain, leading to a nonlinear relationship between the permittivity distribution and the measured inter-electrode capacitances.

\subsection{Geometric Parameterization of Flow Patterns}
The dataset covers four typical ECT two-phase flow patterns: \emph{single-bar}, \emph{two-bar}, \emph{annular}, and \emph{stratified}. Each family is generated by stochastic geometric parameterization to increase morphological diversity while preserving physical interpretability. The center of a circular inclusion is denoted by $\mathbf{c}=(x,y)$.

The four flow patterns are defined as follows.
\begin{itemize}
    \item \textbf{Single-bar flow}: one isolated high-permittivity inclusion.
    \item \textbf{Two-bar flow}: two separated high-permittivity inclusions.
    \item \textbf{Annular flow}: a low-permittivity core surrounded by a high-permittivity phase.
    \item \textbf{Stratified flow}: two layers separated by a straight interface.
\end{itemize}

Table~\ref{tab:geom_random} summarizes the randomized variables and sampling rules for the four flow patterns. Inclusion-based patterns are described in center–radius form, whereas the stratified pattern is parameterized by the distance and orientation of the interface.

\begin{table}[H]
\caption{Geometric parameters of four typical flow patterns}
\label{tab:geom_random}
\centering
\footnotesize
\setlength{\tabcolsep}{3pt}
\renewcommand{\arraystretch}{1.08}
\begin{tabular}{>{\raggedright\arraybackslash}m{1.35cm}
                >{\raggedright\arraybackslash}m{1.65cm}
                >{\raggedright\arraybackslash}m{4.35cm}}
\hline
\textbf{Pattern} & \textbf{Variables} & \textbf{Sampling rule} \\
\hline
Single-bar
& $(\mathbf{c},r)$
& $r \in [2,25]$ mm \\

Two-bar
& \shortstack[c]{$(\mathbf{c}_1,r_1)$,\\ $(\mathbf{c}_2,r_2)$}
& \shortstack[l]{$r_1,r_2 \in [2,20]$ mm;\\ $\|\mathbf{c}_1-\mathbf{c}_2\|_2 \geq r_1+r_2$} \\

Annular
& $(\mathbf{c},r)$
& low-permittivity core; $r \in [5,30]$ mm \\

Stratified
& $(d,\alpha)$
& $d \in [0,0.95R_{\mathrm{in}}],\ \alpha \in [0,2\pi)$ \\
\hline
\end{tabular}
\end{table}

Under this parameterization, each flow pattern forms a continuous distribution rather than a discrete set of templates. The final dataset contains 20,000 samples in total, with 5,000 samples generated for each flow pattern.

\subsection{Generation and Characterization of the Proposed Dataset}

All permittivity--potential--capacitance data are generated through a COMSOL--MATLAB workflow based on the finite element method (FEM) for the electrostatic ECT forward problem. Let $\mathbf{r}=(x,y)$ denote the spatial coordinate in the two-dimensional cross-sectional imaging domain, where $x$ and $y$ correspond to the horizontal and vertical coordinates used in the field maps. Following the quasi-static governing equation introduced in Eq.~\eqref{eq:potential_governing}, the electric potential field in the present two-dimensional setting is obtained by solving its Cartesian form:
\begin{equation}
    \frac{\partial}{\partial x}
    \left(
    \epsilon(x,y)\frac{\partial \phi(x,y)}{\partial x}
    \right)
    +
    \frac{\partial}{\partial y}
    \left(
    \epsilon(x,y)\frac{\partial \phi(x,y)}{\partial y}
    \right)
    =0,
    \label{eq:potential_governing_2d}
\end{equation}
where $\phi(x,y)$ denotes the electric potential and $\epsilon(x,y)$ is the spatially varying permittivity distribution. Eight excitation conditions are considered, in which one electrode is sequentially driven to 1~V while the remaining electrodes are grounded. The terminal charges extracted from the FEM solution are then assembled into a 28-dimensional non-redundant capacitance vector.

The FEM solutions are further organized into spatially registered data arrays. In addition to the capacitance measurements, the permittivity distribution and the excitation-wise electric potential fields are interpolated from the irregular FEM mesh onto a regular Cartesian grid. The released raw dataset retains these quantities on a $78 \times 78$ grid covering the circumscribed square of the outer sensor domain, i.e., the square enclosing the circle of radius 42~mm. The physical coordinates of all grid centers are also provided, so that each field value is explicitly associated with its spatial location.

For geometric interpretation and reconstruction evaluation, the central $64 \times 64$ subgrid corresponds to the circumscribed square of the inner sensing region with radius 35~mm. Only the pixels located inside this inner circular domain are physically effective, yielding 3228 valid pixels in total. Thus, the raw dataset preserves the full $78 \times 78$ field representation, while the $64 \times 64$ subgrid and its 3228 in-circle pixels define the conventional ECT reconstruction domain.

Importantly, the dataset stores all simulation outputs in raw form rather than in a pre-normalized form. To facilitate downstream use, FEM results for the empty-pipe and full-pipe states are also provided, from which users may perform the standard ECT capacitance normalization:
\begin{equation}
    \tilde{\mathbf{C}}=
    \frac{\mathbf{C}-\mathbf{C}_{\mathrm{empty}}}
    {\mathbf{C}_{\mathrm{full}}-\mathbf{C}_{\mathrm{empty}}},
\end{equation}
where $\mathbf{C}$ is the raw 28-dimensional capacitance vector, and $\mathbf{C}_{\mathrm{empty}}$ and $\mathbf{C}_{\mathrm{full}}$ denote the corresponding empty- and full-pipe references. The data structure of one raw ECT sample is summarized in table~\ref{tab:data_format}.

\begin{table}[H]
\caption{Data structure of one raw ECT sample.}
\label{tab:data_format}
\centering
\footnotesize
\renewcommand{\arraystretch}{1.10}
\begin{tabular}{lll}
\hline
\textbf{Variable} & \textbf{Shape} & \textbf{Content} \\
\hline
$\mathbf{E}$ (\textit{Data\_Eps}) & $78 \times 78$ & permittivity distribution \\
$\mathbf{\Phi}$ (\textit{Data\_Pot}) & $78 \times 78 \times 8$ & 8 excitation-wise potential maps \\
$\mathbf{C}$ (\textit{Data\_Cap}) & $28$ & capacitance vector \\
$\mathbf{X}, \mathbf{Y}$ (\textit{X\_grid}, \textit{Y\_grid}) & $78 \times 78$ each & grid-center coordinates \\
\hline
\end{tabular}
\end{table}

\begin{figure}[H]
    \centering
    \includegraphics[width=0.9\textwidth]{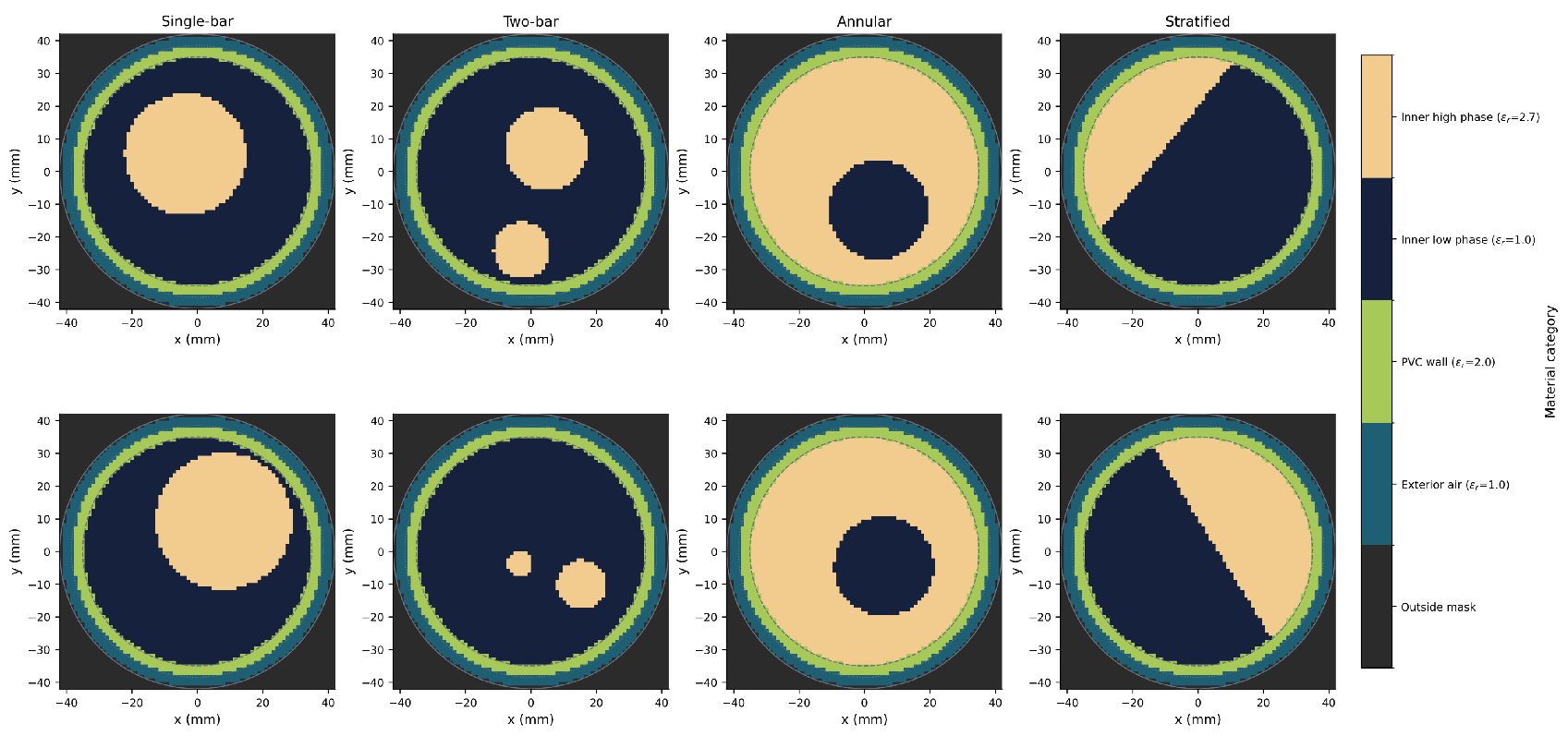}
    \caption{Representative permittivity distributions of the proposed electric potential-augmented ECT dataset. The images depict two typical samples for each of the four flow patterns, namely single-bar, two-bar, annular, and stratified flows, shown on the full $78\times78$ computational grid.}
    \label{fig:dataset_examples_permittivity}
\end{figure}

The structural diversity of the generated samples is illustrated in figure~\ref{fig:dataset_examples_permittivity}. Representative permittivity maps are shown for the four typical flow patterns, namely single-bar, two-bar, annular, and stratified flows. Two samples are displayed for each flow pattern on the full $78 \times 78$ computational grid.

\begin{figure}[H]
    \centering
    \includegraphics[width=0.8\textwidth]{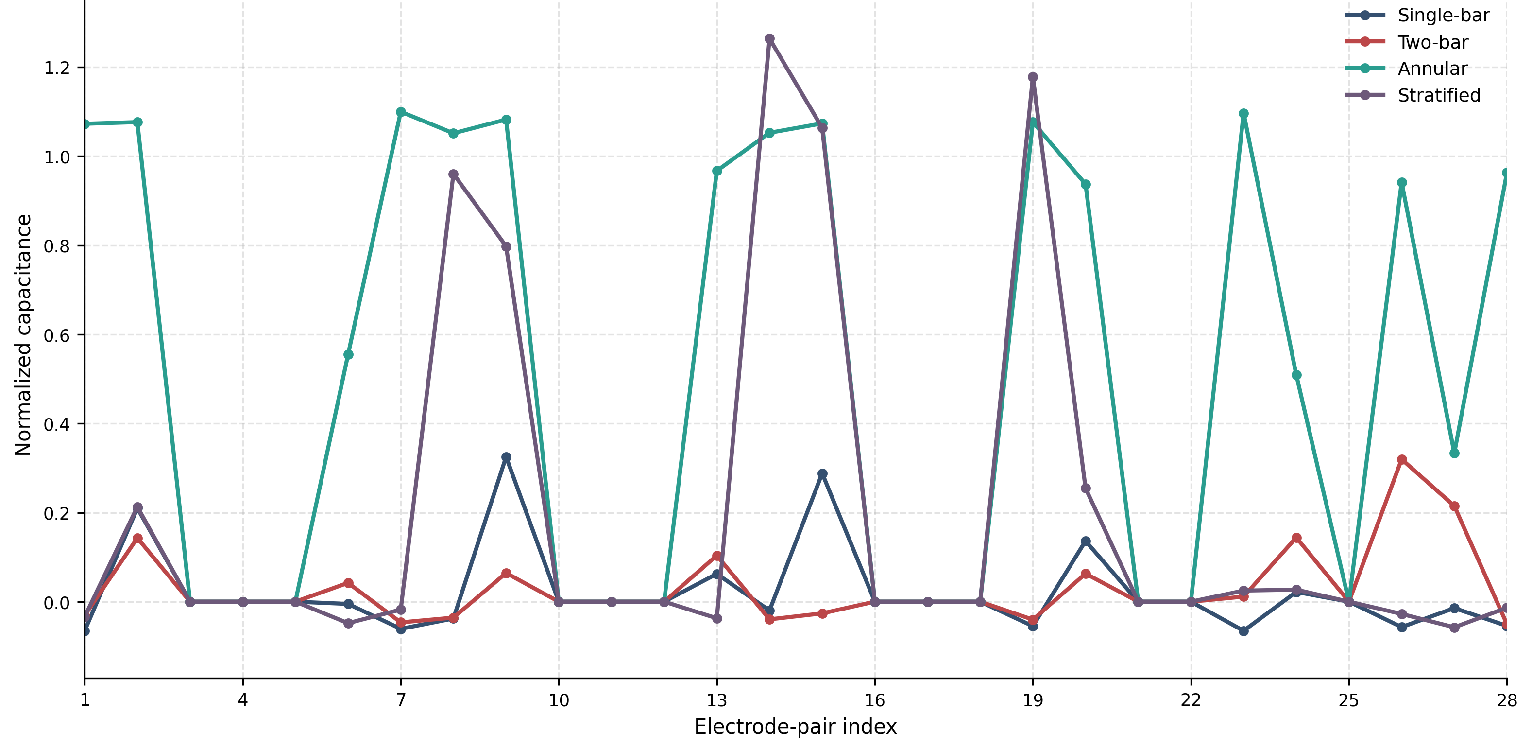}
    \caption{Normalized capacitance vectors of the four representative samples shown in the first row of figure~\ref{fig:dataset_examples_permittivity}.}
    \label{fig:dataset_examples_capacitance}
\end{figure}

Figure~\ref{fig:dataset_examples_capacitance} shows the normalized 28-dimensional capacitance vectors obtained from four representative permittivity distributions, corresponding to the samples displayed in the first row of figure~\ref{fig:dataset_examples_permittivity}. Along each curve, the 28 points represent the normalized inter-electrode capacitance measurements from the non-redundant electrode pairs of the eight-electrode sensor. The differences among the curves indicate that different permittivity distributions produce distinct capacitance measurements under the same sensing configuration.

\begin{figure}[H]
    \centering
    \includegraphics[width=\textwidth]{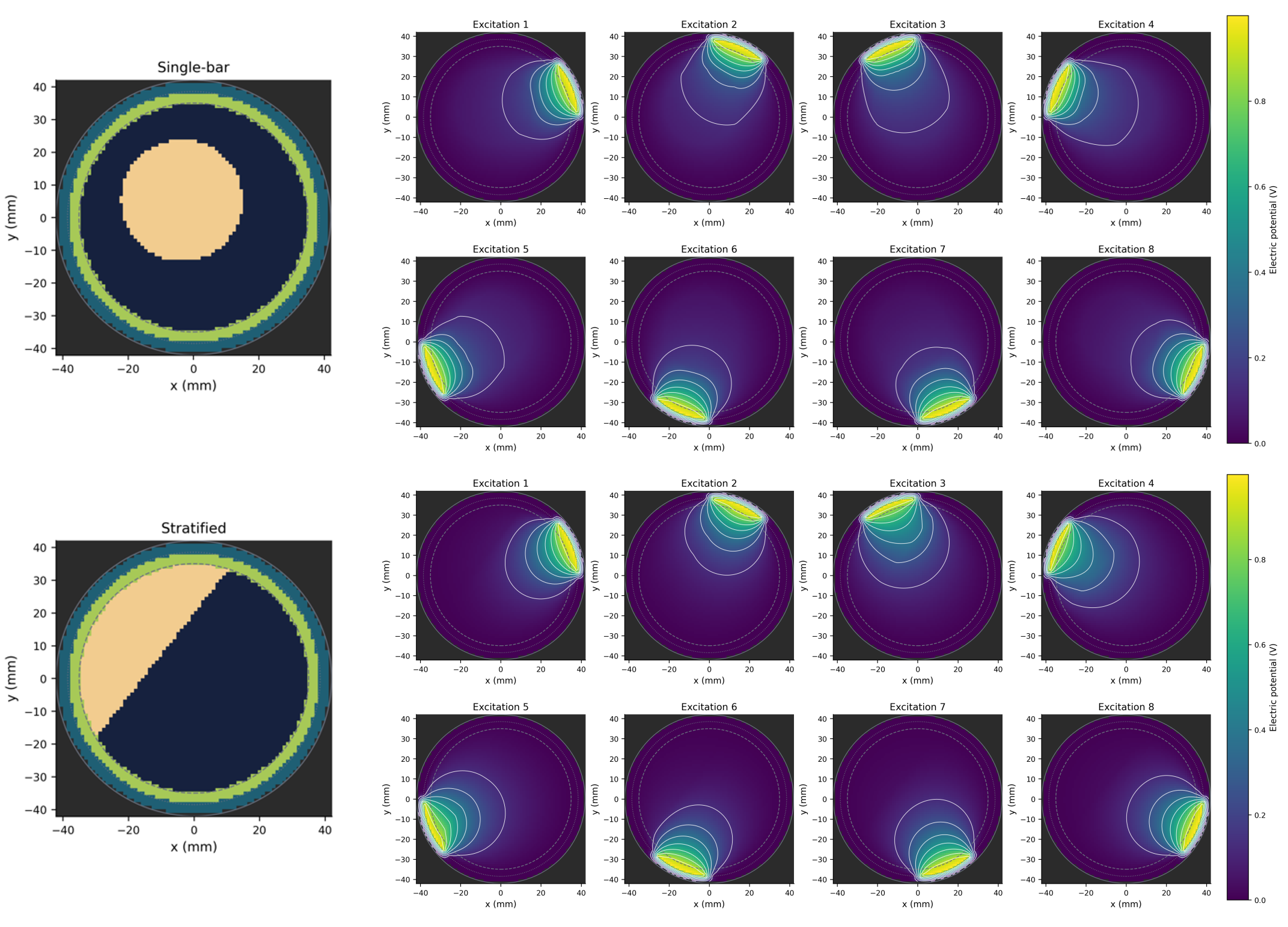}
    \caption{Excitation-wise electric potential fields with equipotential contours for the single-bar and stratified samples shown in the first and fourth columns of the first row of figure~\ref{fig:dataset_examples_permittivity}.}
    \label{fig:dataset_examples_coupled}
\end{figure}

Figure~\ref{fig:dataset_examples_coupled} presents the excitation-wise electric potential fields for two representative flow patterns, namely single-bar and stratified flow, corresponding to the first and fourth samples in the first row of figure~\ref{fig:dataset_examples_permittivity}. The left panel shows the permittivity distribution, and the right panels show the eight potential maps under sequential electrode excitations with equipotential contours overlaid. The local deflection and non-uniform spacing of the contours near high-permittivity regions or phase interfaces indicate that the spatially varying permittivity modifies the potential gradient, illustrating the soft-field effect of ECT.

\begin{figure}[H]
    \centering
    \includegraphics[width=0.9\textwidth]{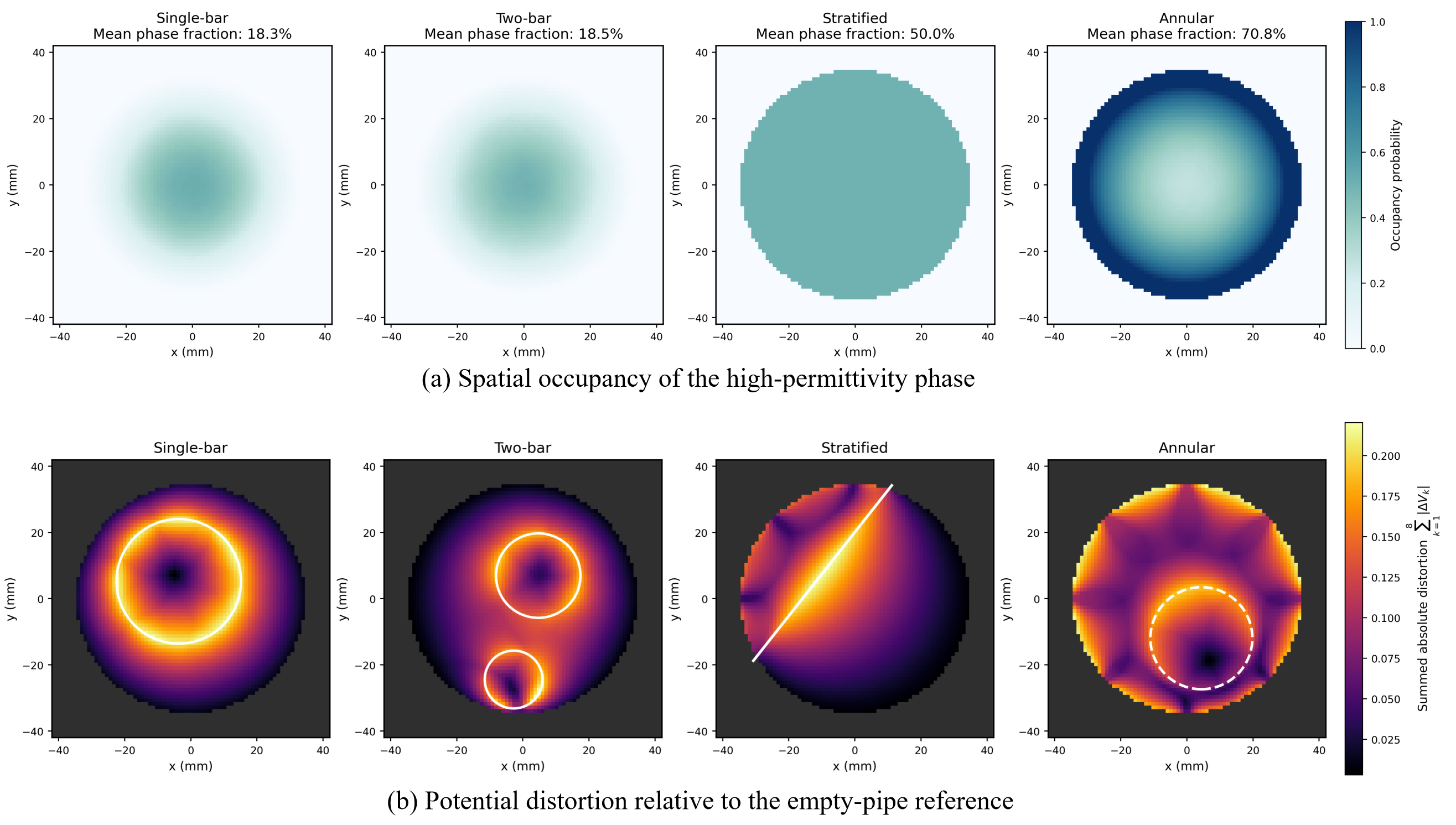}
    \caption{Statistical and field-response diagnostics of the proposed dataset. (a) Average spatial occupancy of the high-permittivity phase for the four flow patterns, summarizing the statistical coverage of the stochastic geometry generators. (b) Potential distortion relative to the empty-pipe reference for representative samples from the four flow patterns. Bright white lines are overlaid to show the outlines of the underlying flow geometries, with a dashed circle used for the annular pattern.}
    \label{fig:dataset_diagnostics}
\end{figure}

To further characterize the dataset beyond individual examples, we examine both the statistical coverage of the generated geometries and the field-response behavior induced by the permittivity distributions. Figure~\ref{fig:dataset_diagnostics}(a) shows the average occupancy of the high-permittivity phase over 5,000 samples for each flow pattern. The resulting maps indicate that the stochastic geometry generators provide broad and stable coverage of the admissible configurations. The variation in color intensity across flow patterns is expected, since the maps represent average phase occupancy and therefore also reflect differences in the mean phase fraction of the corresponding flow patterns.

Figure~\ref{fig:dataset_diagnostics}(b) shows the summed absolute potential deviation from the empty-pipe reference over the eight excitation conditions. The distortion map is computed from the excitation-wise potential distortions and highlights regions where the internal electric potential field is most affected by the permittivity distribution. Regions of relatively large response often appear near phase boundaries, while the affected area generally extends beyond the boundaries themselves. This suggests that the potential distortion is associated with the soft-field effect induced by the location and geometry of the permittivity distribution, indicating that the electric potential field can serve as a physically meaningful intermediate representation for ECT modeling and permittivity prediction.

\section{Experiments on the Effectiveness of Electric Potential Maps}

\subsection{Experimental Protocol and Data Split}

To verify whether explicit electric potential maps are beneficial for ECT machine learning, we evaluate both the forward and inverse problems of ECT under a unified experimental protocol. In the forward problem, the model predicts the normalized 28-dimensional capacitance vector from the permittivity distribution. In the inverse problem, the model reconstructs the permittivity distribution from the normalized 28-dimensional capacitance vector. This design allows the role of electric potential maps to be examined in both directions of the ECT mapping.

In all experiments, the raw FEM outputs are cropped to the central $64 \times 64$ window, and only the inner circular sensing region is retained by a binary mask. The permittivity distribution is linearly normalized to $[0,1]$, the capacitance vector is normalized using the empty- and full-pipe references, and the potential maps are standardized channel-wise using training-set statistics.

The experimental split is designed to assess both in-distribution performance and cross-pattern generalization, as summarized in table~\ref{tab:benchmark_split}. For both the forward and inverse ECT problems, the training set is constructed from the 
\emph{single-bar}, \emph{stratified}, and \emph{annular} flow patterns, with 4,000 samples 
from each pattern, yielding 12,000 training samples in total. The remaining validation and 
test subsets are sample-disjoint from the training set and from each other. The \emph{two-bar} 
flow pattern is reserved for held-out evaluation, with 500 samples used for validation and 
another 500 non-overlapping samples used for out-of-distribution (OOD) testing. In addition, 
an in-distribution (IID) test set contains 1,500 held-out samples from the seen flow patterns, 
including 500 samples from each of the \emph{single-bar}, \emph{stratified}, and 
\emph{annular} patterns.

\begin{table}[H]
\caption{Benchmark split for forward and inverse ECT validation. All subsets are mutually sample-disjoint.}
\label{tab:benchmark_split}
\centering
\footnotesize
\renewcommand{\arraystretch}{1.10}
\begin{tabular}{lll}
\hline
\textbf{Subset} & \textbf{Flow patterns} & \textbf{\# Samples} \\
\hline
Training 
& single-bar, stratified, annular 
& 12,000 \\

Validation 
& two-bar 
& 500 \\

OOD test 
& two-bar 
& 500 \\

IID test 
& single-bar, stratified, annular 
& 1,500 \\
\hline
\end{tabular}
\end{table}

The two-bar pattern is selected as the OOD case because it contains multiple separated inclusions that are absent from the training patterns, while sharing the same sensor configuration and physical model. This setting evaluates cross-pattern generalization from relatively simple or structured phase distributions to a more complex multi-object geometry.

\subsection{Model Variants and Training Strategy}

\begin{figure}[H]
    \centering
    \includegraphics[width=1\textwidth]{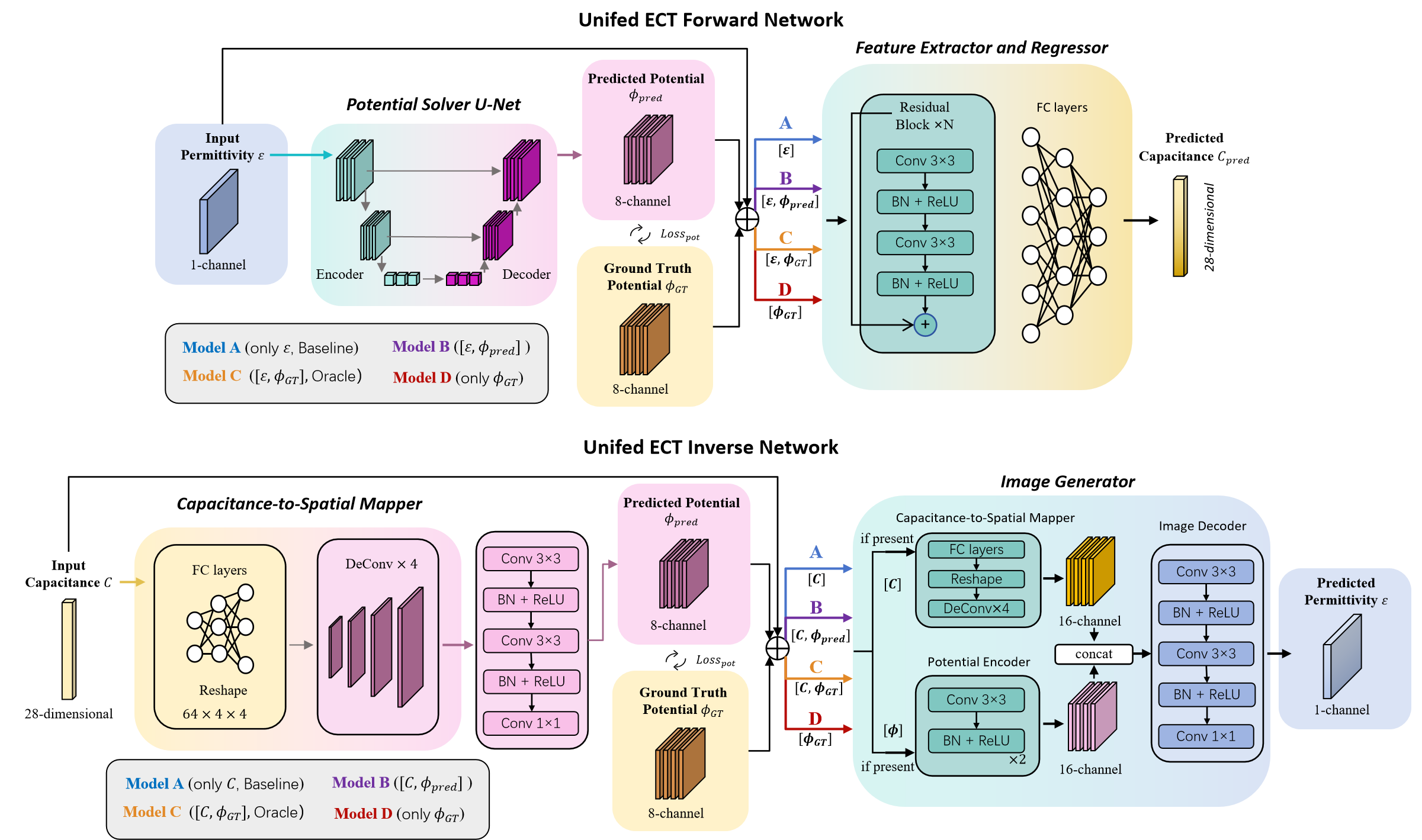}
    \caption{Network architectures used for the potential-effectiveness experiments on forward and inverse ECT problems. The upper part shows the forward models, which predict the normalized 28-dimensional capacitance vector from the permittivity distribution. The lower part shows the inverse models, which reconstruct the permittivity distribution from the normalized capacitance vector.}
    \label{fig:ect_model_architectures}
\end{figure}
To isolate the role of electric potential maps, four model variants are evaluated for both forward and inverse ECT problems. These variants are defined as follows:
\begin{itemize}
    \item \textbf{Model A (baseline):} the task-input baseline using only the original task-specific input.
    \item \textbf{Model B (potential-augmented):} the practical potential-augmented model using predicted potential maps together with the task-specific input.
    \item \textbf{Model C (oracle reference):} an oracle reference that replaces the predicted potential maps in Model~B with ground-truth potential maps.
    \item \textbf{Model D (potential-only diagnostic):} a diagnostic setting using only ground-truth potential maps without the task-specific input.
\end{itemize}

These model settings are summarized in table~\ref{tab:model_variants}. The comparison between Models~A and B evaluates the practical benefit of incorporating predicted potential maps into the task-specific input. Model~C provides an upper-bound reference using ground-truth potential maps, whereas Model~D evaluates the target relevance of the potential maps alone.

\begin{table}[H]
\caption{Definition of the model variants used in the potential-effectiveness experiments.}
\label{tab:model_variants}
\centering
\small
\begin{tabular}{lll}
\hline
Model & Role & Input setting \\
\hline
A & Baseline & Task-specific input only \\
B & Potential-augmented & Task-specific input + predicted potential maps \\
C & Oracle reference & Task-specific input + ground-truth potential maps \\
D & Potential-only diagnostic & Ground-truth potential maps only \\
\hline
\end{tabular}
\end{table}

Here, the task-specific input denotes the permittivity distribution for the forward problem and the normalized capacitance vector for the inverse problem. Figure~\ref{fig:ect_model_architectures} summarizes the corresponding model architectures. For the forward problem, the network predicts the 28-dimensional capacitance vector from the permittivity distribution. For the inverse problem, the network reconstructs the permittivity distribution from the capacitance vector. The four model variants are implemented consistently in the two tasks so that the role of electric potential maps can be evaluated under a unified comparison framework.

Model~B is the central practical model because it introduces predicted potential maps into the learning pipeline without requiring ground-truth potential information at inference time. In the forward problem, Model~B first predicts eight excitation-wise potential maps from the input permittivity distribution and then fuses the predicted potential maps with the permittivity representation for capacitance regression. In the inverse problem, Model~B first estimates the excitation-wise potential maps from the capacitance vector and then combines the predicted potential features with capacitance-derived spatial features for permittivity reconstruction. Model~C replaces the predicted potential maps in Model~B with ground-truth potential maps, providing an upper-bound reference for accurate field information. Model~D removes the task-specific input and uses only ground-truth potential maps, serving as a diagnostic model for assessing the target relevance of the potential field itself.

Because ground-truth potential maps are unavailable during practical inference, Model B must
first estimate them from the available task-specific input before using them for downstream
prediction. The additional potential-estimation module is therefore part of the tested
potential-augmented pipeline, rather than an auxiliary enlargement of the baseline backbone.
Accordingly, the A--B comparison is interpreted as a pipeline-level ablation of estimated
potential-field information: Model A uses only the conventional task-specific input, whereas
Model B makes an estimated latent field representation available for the same ECT task. Models C
and D then contextualize this comparison by indicating the value of accurate potential maps and
the standalone information content of the potential field.

The current implementation adopts a two-stage training strategy for Model~B in both ECT problems. In the forward problem, the potential predictor is first trained with supervised potential reconstruction and a PDE-inspired regularization term; after that, the predictor is frozen and the downstream capacitance regressor is trained using the fused representation. In the inverse problem, the capacitance-to-potential generator is first trained and then frozen, while the fusion encoder and image decoder are optimized for permittivity reconstruction. This design allows us to directly examine whether learned electric potential maps can serve as useful intermediate physical representations for ECT forward modeling and image reconstruction.

\subsection{Quantitative Evaluation}

The forward problem is evaluated using MSE and MAE computed on the normalized 28-dimensional capacitance vector, whereas the inverse problem is evaluated using MSE and MAE computed on the reconstructed permittivity distribution within the valid circular sensing region. Thus, the inverse-problem errors are restricted to the physically meaningful imaging domain. The quantitative results for both forward and inverse ECT problems are reported in table~\ref{tab:potential_results}.

\begin{table}[H]
\caption{Quantitative results of the potential-effectiveness experiments on forward and inverse problems of ECT.}
\label{tab:potential_results}
\centering
\small
\setlength{\tabcolsep}{5pt}
\begin{tabular}{llcccc}
\hline
Task & Model & IID MSE $\downarrow$ & IID MAE $\downarrow$ & OOD MSE $\downarrow$ & OOD MAE $\downarrow$ \\
\hline
Forward & A & $2.98\times10^{-5}$ & $4.21\times10^{-3}$ & $3.53\times10^{-4}$ & $1.13\times10^{-2}$ \\
Forward & B & $2.06\times10^{-5}$ & $3.36\times10^{-3}$ & $2.76\times10^{-4}$ & $1.01\times10^{-2}$ \\
Forward & C & $1.78\times10^{-5}$ & $3.18\times10^{-3}$ & $1.94\times10^{-4}$ & $9.00\times10^{-3}$ \\
Forward & D & $6.78\times10^{-5}$ & $6.25\times10^{-3}$ & $4.88\times10^{-4}$ & $1.46\times10^{-2}$ \\
\hline
Inverse & A & $4.44\times10^{-3}$ & $1.30\times10^{-2}$ & $2.21\times10^{-2}$ & $3.29\times10^{-2}$ \\
Inverse & B & $3.45\times10^{-3}$ & $1.16\times10^{-2}$ & $2.17\times10^{-2}$ & $3.27\times10^{-2}$ \\
Inverse & C & $2.81\times10^{-4}$ & $2.73\times10^{-3}$ & $9.64\times10^{-4}$ & $3.17\times10^{-3}$ \\
Inverse & D & $1.04\times10^{-3}$ & $8.68\times10^{-3}$ & $1.60\times10^{-3}$ & $1.05\times10^{-2}$ \\
\hline
\end{tabular}
\end{table}

The most practically relevant comparison is between Model~A and Model~B, because both models use deployable inputs, while Model~B further incorporates predicted potential maps. For the forward problem, Model~B consistently outperforms Model~A, reducing IID MSE/MAE by approximately 30.9\%/20.2\% and OOD MSE/MAE by approximately 21.8\%/10.6\%. This improvement indicates that predicted potential maps provide useful intermediate electrostatic-field information for the mapping from permittivity distribution to capacitance measurements.

For the inverse problem, Model~B also improves over Model~A in the IID setting, reducing MSE/MAE by approximately 22.3\%/10.8\%. However, the improvement becomes much smaller under the OOD two-bar setting, with only about 1.8\% reduction in MSE and 0.6\% reduction in MAE. This contrast suggests that predicted potential maps can assist reconstruction when the testing distribution is close to the training patterns, but their effectiveness is limited when reliable field information must be inferred from sparse capacitance measurements for unseen multi-object geometries.

Models~C and D provide further diagnostic evidence. Model~C uses ground-truth potential maps together with the task-specific input and therefore represents an upper-bound reference for accurate potential-field information. Its strong performance, especially in the inverse problem, indicates that the electric potential field contains rich spatial information related to the underlying permittivity distribution. Model~D uses only ground-truth potential maps. Its weaker forward performance shows that potential maps alone cannot replace the original permittivity input for capacitance prediction, whereas its strong inverse performance suggests that full-field potential maps are highly informative for recovering material structure. Therefore, the main value of potential maps lies not in replacing the original ECT input, but in serving as auxiliary latent field information that complements the task-specific representation.

\subsection{Qualitative Evaluation}

To complement the quantitative results in table~\ref{tab:potential_results}, representative visual examples are provided for both inverse reconstruction and forward capacitance prediction. The inverse results compare the reconstructed permittivity distributions of different model variants, whereas the forward results compare the predicted capacitance vectors and their component-wise errors. Together, these visualizations illustrate the sample-level behavior behind the quantitative trends.

\begin{figure}[H]
    \centering
    \includegraphics[width=\textwidth]{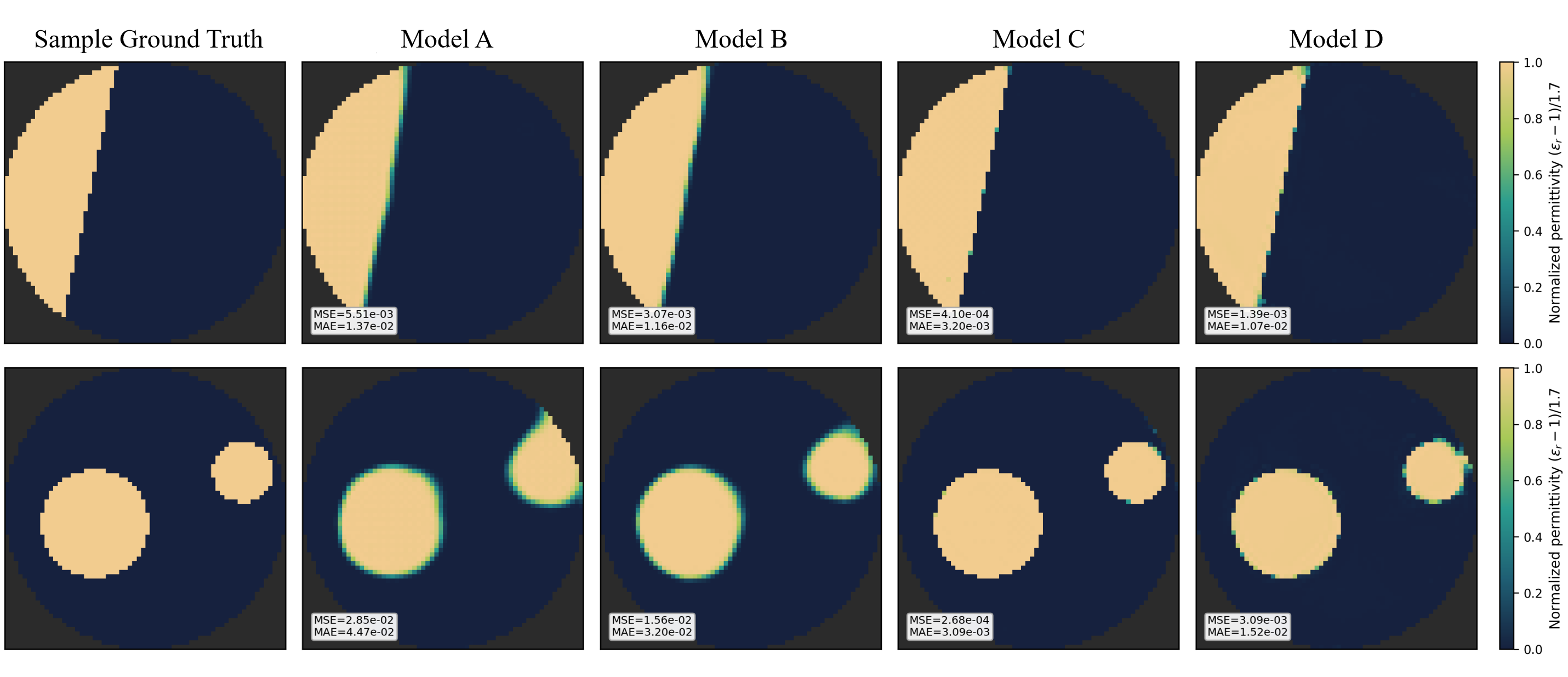}
    \caption{Inverse reconstruction examples for representative stratified and two-bar samples. For each sample, the ground-truth permittivity distribution is compared with the reconstructed results of Models~A--D.}
    \label{fig:inverse_qualitative_iid_ood}
\end{figure}

The representative examples in figure~\ref{fig:inverse_qualitative_iid_ood} show trends that are consistent with the quantitative results. For the stratified sample, Model~B produces a straighter and smoother interface than Model~A and reduces the reconstruction error. For the OOD two-bar sample, Model~B better preserves the separated high-permittivity regions and also achieves lower sample-level MSE and MAE than Model~A. Models~C and D further show that ground-truth potential maps contain strong spatial information about the permittivity distribution; however, since these potential maps are not available in practical ECT measurements, the two models should be interpreted as oracle or diagnostic references.

\begin{figure}[H]
    \centering
    \includegraphics[width=\textwidth]{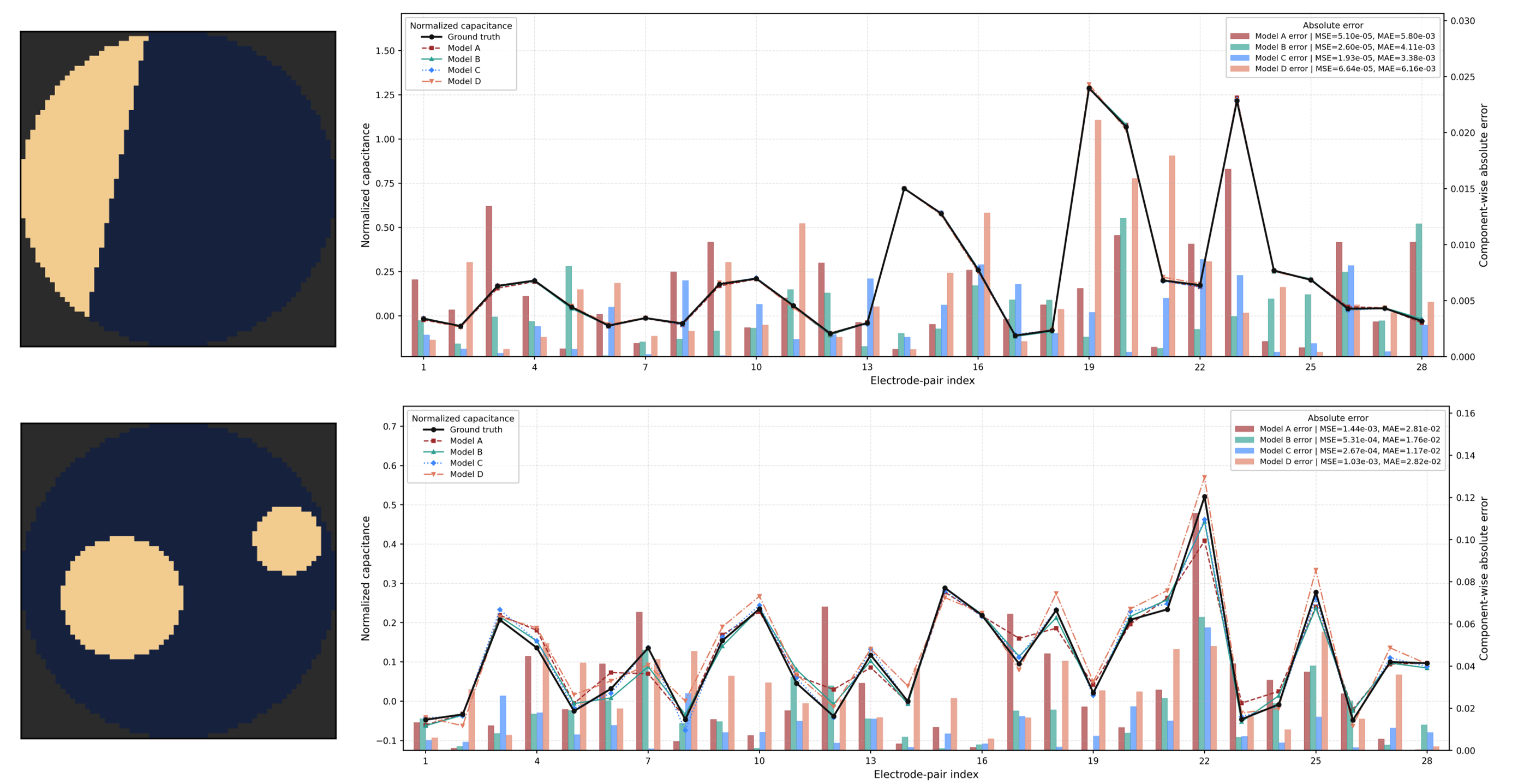}
    \caption{Forward prediction results for representative stratified and OOD two-bar samples. For each sample, the permittivity distribution is shown on the left, while the normalized capacitance-vector comparison and the component-wise absolute errors are overlaid on the right. Line plots denote the ground-truth and predicted normalized capacitance vectors, and grouped bars denote the absolute errors over the 28 non-redundant electrode-pair measurements.}
    \label{fig:forward_qualitative_ood}
\end{figure}

Figure~\ref{fig:forward_qualitative_ood} shows the forward prediction results for representative IID stratified and OOD two-bar samples. The line plots denote the ground-truth and predicted normalized capacitance vectors, whereas the grouped bars denote the corresponding component-wise absolute errors. For the IID stratified sample, all models produce relatively small errors, but Model~B still follows the ground-truth capacitance vector more closely than Model~A and gives lower sample-level MSE and MAE. For the OOD two-bar sample, the overall errors are larger, and the differences among the models become more pronounced; in this more challenging case, Model~B also reduces the sample-level MSE and MAE compared with Model~A. The component-wise error patterns further indicate that the improvement varies across electrode pairs, reflecting the non-uniform sensitivity of ECT capacitance measurements to the spatial permittivity distribution.

Overall, the quantitative and qualitative results consistently show that electric potential maps are useful for both ECT forward modeling and inverse reconstruction. The practical gain of Model~B demonstrates the benefit of incorporating predicted potential maps, while the remaining gap between Models~B and C, especially in the OOD inverse problem, highlights the importance of improving potential-field estimation for complex unseen flow geometries.

\section{Conclusion and Outlook}
This paper presents an electric potential-augmented benchmark dataset for electrical capacitance tomography (ECT). The dataset contains 20,000 randomized samples from four typical flow patterns and, for each sample, jointly provides the permittivity distribution depicted as an image, eight excitation-wise potential maps, and the corresponding 28-dimensional capacitance vector. Empty-pipe and full-pipe reference states are also included to support data normalization and subsequent ECT studies.

Based on this dataset, we conducted systematic forward and inverse ECT experiments to verify the effectiveness of electric potential maps. The results show that latent field information can effectively enhance ECT modeling and image reconstruction.

The current study is limited to two-dimensional FEM simulation, a single eight-electrode sensor geometry, and a single permittivity contrast pair. Future extensions may include more realistic measurement-noise and uncertainty models, experimental ECT measurements, multiple contrast settings, and three-dimensional sensor configurations. More broadly, the proposed benchmark suggests a general route toward physics-guided machine learning of ECT image reconstruction through latent field representations.

\data{The data that support the findings of this study are openly available in Zenodo at doi:10.5281/zenodo.20495398.}

\section*{Funding}
This study received no specific funding.

\section*{Conflict of interest}
The authors declare no conflicts of interest.

\end{document}